\documentclass[runningheads]{llncs}
\usepackage{graphicx}
\usepackage{amsmath,amssymb} 
\usepackage{color}
\usepackage[width=122mm,left=12mm,paperwidth=146mm,height=193mm,top=12mm,paperheight=217mm]{geometry}
\begin{document}
\pagestyle{headings}
\mainmatter
\def\ECCV16SubNumber{***}  

\title{Does V-NIR based Image Enhancement Come with Better Features?} 

\titlerunning{Does V-NIR based Image Enhancement Come with Better Features?}

\authorrunning{Sharma and Van Gool}

\author{Vivek Sharma$^{\diamond}$ and Luc Van Gool$^{\diamond \mp}$}
\institute{$^{\diamond}$ESAT-PSI, KU Leuven, and 
$^{\mp}$CVL, ETH Z\"{u}rich.\\
        {\tt\small \{vivek.sharma,luc.vangool\}@esat.kuleuven.be}}

\maketitle

\begin{abstract}
Image enhancement using the visible (V) and near-infrared (NIR) usually enhances useful image details. The enhanced images are evaluated by observers perception, instead of quantitative feature evaluation. Thus, can we say that these enhanced images using NIR information has better features in comparison to the computed features in the Red, Green, and Blue color channels directly?   In this work, we present a new method to enhance the visible images using NIR information via edge-preserving filters, and also investigate which method  performs best  from a image features standpoint. We then show that our proposed enhancement method produces more stable features than the existing state-of-the-art methods.
\end{abstract}

\section{Introduction}
Image enhancement using visible and near-infrared (NIR) images is used regularly for several applications, such as aerial or landscape photography, dehazing~\cite{dehazing}, tone mapping~\cite{ep}, biometrics~\cite{raghu} and  more. Visible images enhanced using near-infrared part of electromagnetic spectrum enhances the contrast, details, and produces more vivid colors that are more pleasant to observers perception.  

Over the last two decades, several image enhancement approaches have been proposed. For combining NIR information with RGB images, the NIR channel is combined as either color, luminance or frequency counterpart. This combination is achieved using linear (Laplacian pyramid) or non-linear (anisotropic diffusion, robust smoothing, weighted least squares, and bilateral filtering) filters. Each of the filtering technique comes at some expense, such as high computational time, more artifacts, high  noise level, inability to preserve edges and shape details, and more. All these shortcomings add up to undesired loss of image features, however visually these images may appear very pleasant. To this end, we propose to combine these filters in a meaningful way, such that a minimum loss of visual pleasantness or information content is attained. 

In this work, we focus on two recently emerged as \textit{de facto} tools for edge-preserving filters: bilateral filter (BF)~\cite{tomasi,fbf} and weighted least squares optimization framework (WLS)~\cite{ep}. We propose to combine the base and detail layers (i.e. low and high frequency decompositions) for a pair of visible and NIR images with the WLS and BF filters. As we combine BF and WLS filters, we call our method BFWLS.

The rest of the paper is organized as follows. In Section~\ref{sec:background}, we discuss the related work, and Section~\ref{sec:approach} describes our proposed method. Experiments are given in Section~\ref{sec:experiments}, and Finally, the conclusions are drawn in Section~\ref{sec:conclusion}.

\section{Background} \label{sec:background}

Recently, several approaches have been proposed for image fusion using visible and near-infrared images. They make use of BF and WLS filters to combine the color, luminance or detail layers of NIR image to counterpart the visible image for enhancing the visible images~\cite{coloring}. Such as, Fredembach et al.~\cite{vnir} make use of the BF proposed by Tomasi et al.~\cite{tomasi} to decompose the  V-NIR images into base (low frequency component) and detail (high frequency component) layers, and then swapped the detail layer layer of NIR with the ones of the RGB image for realistic skin smoothing.  Similarly, Schaul et al.~\cite{dehazing}  improve the contrast of the haze-degraded color images by transforming the visible and NIR images into their multiresolution decomposition using WLS filter.

Some other notable work where they use the same idea of image fusion, but exploiting the RGB channels only are like ``Fast Bilateral Filtering" by Durand et al.~\cite{fbf} where they reduce the contrast of the high-dynamic range images to display it on low-dynamic range media using  BF. Similarly,  Farbman et al.~\cite{ep} combines details of RGB channels at various scales using WLS multi-scale image decompositions for tone mapping and detail enhancement.  We compare against all of these image fusion approaches in our experimental section.

\section{Proposed Approach: BFWLS} \label{sec:approach}

\begin{figure}[t]
\centering
{\includegraphics[width=0.90\columnwidth]{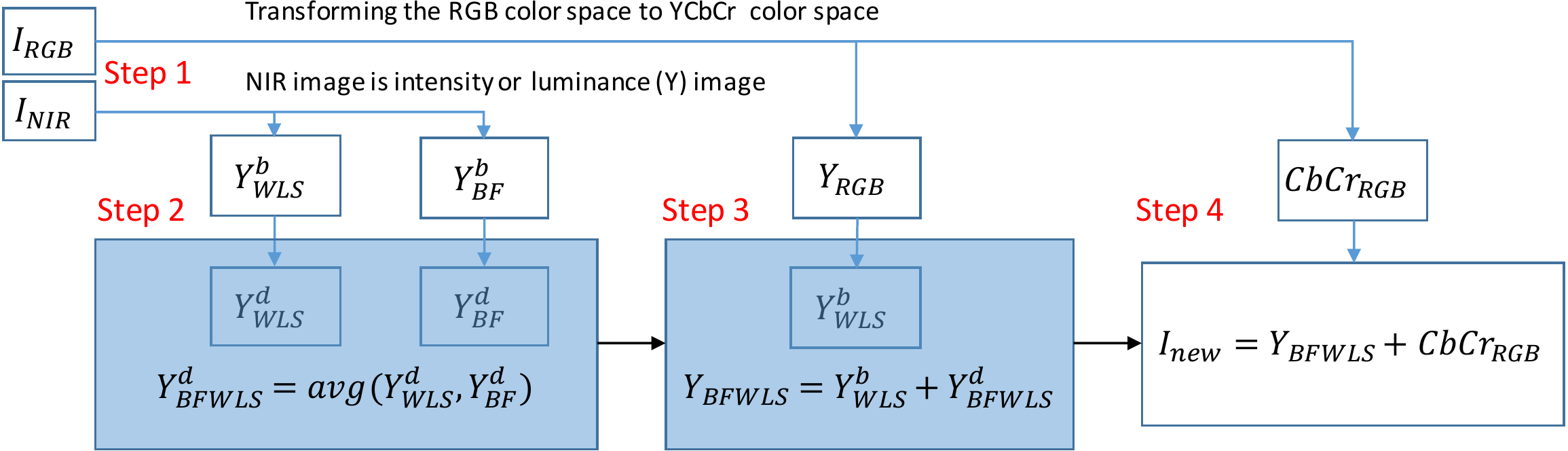} 
} 
\caption{\textbf{BFWLS:} For an input pair of images (I$_{RGB}$, I$_{NIR}$), the intermediate base (b) and detail (d) layers are obtained first (Step 1) for both images using BF and WLS filters. In the Step 2-3, the new fused luminance images is obtained, and it is then combined with chrominance of RGB image (Step 4), to construct the new image. 
} \label{fig:approach}\vspace*{-0.5cm}
\end{figure}

Fig.~\ref{fig:approach} illustrates the entire procedure of our proposed approach. To fuse the visible and NIR images, we first transform the RGB colourspace into the luminance-chrominance colourspace, where the NIR image is already the luminance image~\cite{coloring}. The chrominance is not used in the fusion algorithm, but simply recombined in the final fused image. 

The BF and WLS filters decompose an image into base and detail images. The detail images are obtained by simply subtracting the original image minus the base image. The base image comprises low frequency content with general appearance of the image over smooth areas, while the detail layer comprises of high frequency contents	with edges   and   sharp   transition (e.g.  noise). 

We apply WLS-based and BF-based decomposition of the NIR image for extraction of base and detail images. We retain, for each pixel the average values of the detail-WLS and detail-BF images. We chose $\lambda=0.125$ and $c=1.2$ for WLS~\cite{ep}; and $edge_{min}=0.2$, $edge_{max}=1.0$, $\sigma_{spatial}=42.43$ and $\sigma_{range}=0.1$ for BF~\cite{fbf} in our experiments.

The fusion criteria is based on the following observations: the BF filter preserves edges and can extract details at a fine spatial scale, but lacks the ability to extract details at arbitrary scales. Where as, WLS filter is very good at preserving  fine and coarse details at arbitrary scales. Taking an average between two, - allows to retain the details from both, - moderately boosts the details, and - we also found that the hidden details appeared in this way. 

As a fusion criterion, we also tried to retain the maximum values between the two, but we found  that fusing detail layers in this way suffered from an undesired loss of important information content. 

The base layer of RGB image contains low luminance information as perceived by humans visual system, thus the NIR base layer is discarded.  We combine the fused detail layer of NIR image with the base layer of RGB image obtained using WLS, to obtain the new luminance image. This new luminance image enhances the original image's contrast and details, and it is then combined with chrominance of RGB image to reconstruct the final image.

\begin{table*}[t] 
\begin{center}
\resizebox{12cm}{!} {
\begin{tabular}{| c | c |  c |  c | c | c |  c |  }
 \hline
 Metric &  Fredembach  & WLS (replaced BF & Schaul & Farbman  & BFWLS-& BFWLS-\\
 (Average)& et al.~\cite{vnir} & by WLS in~\cite{vnir}) \textbf{(ours)}  & et al.~\cite{dehazing}  & et al.~\cite{ep} & Max \textbf{(ours)}& Avg\textbf{(ours)}\\
\hline
Rel. Change (\%) &4.31&3.82&-1.11&4.92&3.29&\textbf{8.78}\\
\hline
Time (Sec) &\textbf{0.20}&6.54&38.20&6.65&6.39&6.59\\
\hline
PSNR &32.36&\textbf{33.45}&22.83&13.28&30.28&32.69\\
\hline
MSE  ($10^{-4}$)&6.76&\textbf{6.42}&61&637&10&6.45\\
\hline
\end{tabular}}
\end{center}
\caption{Comparison of BFWLS with the other methods. For Farbman et al.~\cite{ep}, we replace the luminance of RGB by NIR. The image is of resolution $1024\times680$ pixels.}\vspace{-0.5cm}
\label{table:results}
\end{table*}

\section{Experiments}\label{sec:experiments}
We evaluate our proposed method on 477 pairs of images from RGB-NIR Scene Dataset~\cite{dataset}.   We evaluate the features quality by analysing the amount of original features that remain after applying the transformation to an image. To this end, we apply synthetic transformations: rotation ($45^{\circ}$, $90^{\circ}$ and $180^{\circ}$), and scaling (0.5, 0.75) to each image. Finally the feature matching is done between the original and transformed  image pairs, where  $threshold=1.5$, producing a number of matches. We compare the matches obtained from the fusion algorithm against the matches extracted from RGB images, and report the relative change (in \%) as an evaluation criterion. For feature matching, we use the SIFT implementation from the  Vlfeat library~\cite{vlfeat}. For the SIFT descriptors, we use a bin-size of 8 and step-size of 4.

\begin{figure}[t]
\centering
{\includegraphics[width=0.99\columnwidth]{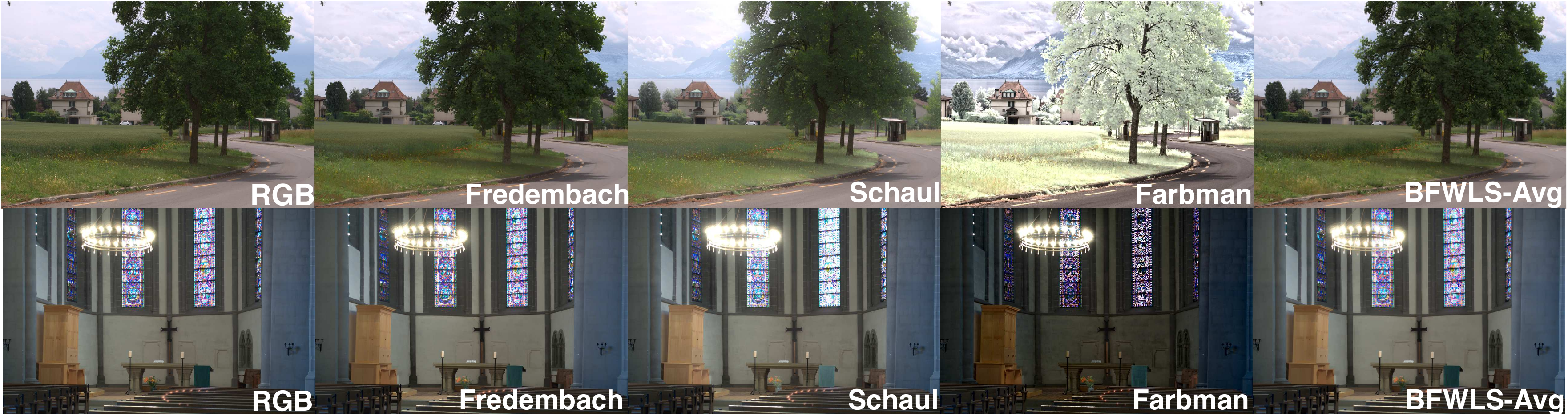} 
} \vspace*{-0.45cm}
\caption{ Comparison of the image enhancement approaches against RGB image, Fredembach et al.~\cite{vnir}, Schaul et al.~\cite{dehazing}, Farbman et al.~\cite{ep}, and our BFWLS-Avg method. } \label{fig:visualresults}\vspace{-0.5cm}
\end{figure}

We can see in Table.~\ref{table:results} that our method has more feature matches over state-of-the-art methods. Note that Schaul et al.~\cite{dehazing} feature matches degrade by 1.11\% against RGB images after the image fusion.   Our enhanced images has better high-frequency details, and further improve the ability to preserve edges because our approach can extract details at fine spatial and arbitrary scales due of combined fusion from WLS and BF filters, shown in Fig.~\ref{fig:visualresults}. Additional qualitative results  and the source code for all the experiments will be available soon at \textit{\url{http://homes.esat.kuleuven.be/$\sim$vsharma/BFWLS.html}}

\section{Conclusion} \label{sec:conclusion}
We present a method to combine visible and near-infrared images using edge-preserving filters: bilateral filter and weighted least square filter. Our method successfully  enhances the visible images using near-infrared information, and also improve the image features quality over the state-of-the-art methods. 

{
\scriptsize
\bibliographystyle{splncs}
\bibliography{egbib}

\begin{thebibliography}{1}

\bibitem{dehazing}
Schaul, L., Fredembach, C., S{\"u}sstrunk, S.:
\newblock Color image dehazing using the near-infrared.
\newblock In: ICIP'09

\bibitem{ep}
Farbman, Z., Fattal, R., Lischinski, D., Szeliski, R.:
\newblock Edge-preserving decompositions for multi-scale tone and detail
  manipulation.
\newblock In: ACM TOG'08

\bibitem{raghu}
Raghavendra, R., Busch, C.:
\newblock Novel image fusion scheme based on dependency measure for robust
  multispectral palmprint recognition.
\newblock IEEE Pattern Recognition'14

\bibitem{tomasi}
Tomasi, C., Manduchi, R.:
\newblock Bilateral filtering for gray and color images.
\newblock In: ICCV'98

\bibitem{fbf}
Durand, F., Dorsey, J.:
\newblock Fast bilateral filtering for the display of high-dynamic-range
  images.
\newblock In: ACM TOG'02

\bibitem{coloring}
Fredembach, C., S{\"u}sstrunk, S.:
\newblock Colouring the near-infrared.
\newblock In: CIC'08

\bibitem{vnir}
Fredembach, C., Barbuscia, N., S{\"u}sstrunk, S.:
\newblock Combining visible and near-infrared images for realistic skin
  smoothing.
\newblock In: CIC'09

\bibitem{dataset}
Brown, M., S{\"u}sstrunk, S.:
\newblock Multi-spectral sift for scene recognition.
\newblock In: CVPR'11

\bibitem{vlfeat}
Vedaldi, A., Fulkerson, B.:
\newblock Vlfeat: An open and portable library of computer vision algorithms.
\newblock In: ACM MM'10

\end{thebibliography}
}

\end{document}